\documentclass{article}
\usepackage{arxiv}
\usepackage[utf8]{inputenc} % allow utf-8 input
\usepackage[T1]{fontenc}    % use 8-bit T1 fonts
\usepackage{hyperref}       % hyperlinks
\usepackage{url}            % simple URL typesetting
\usepackage{booktabs}       % professional-quality tables
\usepackage{amsmath,amssymb,amsfonts,amsthm} % blackboard math symbols
\usepackage{nicefrac}       % compact symbols for 1/2, etc.
\usepackage{microtype}      % microtypography
\usepackage{lipsum}		% Can be removed after putting your text content
\usepackage{graphicx}
\usepackage{natbib}
\usepackage{doi}

\usepackage{algorithmic}
\usepackage{algorithm}
\usepackage{multirow}

\newcommand{\beq}{\begin{equation}}
\newcommand{\eeq}{\end{equation}}

\newcommand{\bp}{\mathbf{p}}

\newcommand{\bv}{\mathbf{v}}
\newcommand{\bw}{\mathbf{w}}
\newcommand{\bx}{\mathbf{x}}
\newcommand{\by}{\mathbf{y}}
\newcommand{\bz}{\mathbf{z}}

\newcommand{\bX}{\mathbf{X}}

\newcommand{\bone}{\mathbf{1}}

\newcommand{\bbeta}{\boldsymbol{\beta}}

\newcommand{\btheta}{\boldsymbol{\theta}}

\newcommand{\bmu}{\boldsymbol{\mu}}

\DeclareMathOperator*{\argmax}{arg\,max}
\DeclareMathOperator*{\argmin}{arg\,min}

\newtheorem{theorem}{Theorem}

\title{Subject-specific Deep Neural Networks 
for Count Data with High-cardinality 
Categorical Features}

\date{}

\author{ 
\href{https://orcid.org/0000-0002-3447-4306}
{\includegraphics[scale=0.06]{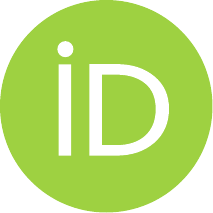}\hspace{1mm}
Hangbin~Lee}\\
Department of Statistics\\
Seoul National University\\
\texttt{hangbin221@gmail.com} \\
\And
\href{https://orcid.org/0000-0001-8434-9506}
{\includegraphics[scale=0.06]{orcid.pdf}\hspace{1mm}
Il~Do~Ha} \\
Department of Statistics\\
Pukyong National University\\
\texttt{idha1353@pknu.ac.kr} \\
\AND
\href{https://orcid.org/0000-0003-1262-5051}
{\includegraphics[scale=0.06]{orcid.pdf}\hspace{1mm}
Changha~Hwang} \\
Department of Statistics\\
Dankook University\\
\texttt{chwang@dankook.ac.kr} \\
\And
\href{https://orcid.org/0000-0001-9820-6434}
{\includegraphics[scale=0.06]{orcid.pdf}\hspace{1mm}
Youngjo~Lee} \\
Department of Statistics\\
Seoul National University\\
\texttt{youngjo@snu.ac.kr} \\
}

\renewcommand{\shorttitle}{Subject-specific DNN for count data}

\begin{document}
\maketitle

\begin{abstract}
There is a growing interest in subject-specific predictions using deep neural networks (DNNs) because real-world data often exhibit correlations, which has been typically overlooked in traditional DNN frameworks. In this paper, we propose a novel hierarchical likelihood learning framework for introducing gamma random effects into the Poisson DNN, so as to improve the prediction performance by capturing both nonlinear effects of input variables and subject-specific cluster effects. The proposed method simultaneously yields maximum likelihood estimators for fixed parameters and best unbiased predictors for random effects by optimizing a single objective function. This approach enables a fast end-to-end algorithm for handling clustered count data, which often involve high-cardinality categorical features. Furthermore, state-of-the-art network architectures can be easily implemented into the proposed h-likelihood framework. As an example, we introduce multi-head attention layer and a sparsemax function, which allows feature selection in high-dimensional settings. To enhance practical performance and learning efficiency, we present an adjustment procedure for prediction of random parameters and a method-of-moments estimator for pretraining of variance component. Various experiential studies and real data analyses confirm the advantages of our proposed methods.
\end{abstract}

\section{Introduction}

Deep neural networks (DNNs), which have been proposed 
to capture the nonlinear relationship 
between input and output variables \citep{lecun15, goodfellow16},
provide outstanding marginal predictions for independent outputs.
However, in practical applications, 
it is common to encounter correlated data
with high-cardinality categorical features, 
which can pose challenges for DNNs.
While the traditional DNN framework overlooks such correlation,
random effect models have emerged in statistics
to make subject-specific predictions for correlated data. 
\citet{lee96} proposed hierarchical generalized linear models (HGLMs), 
which allow the incorporation of random effects 
from an arbitrary conjugate distribution 
of generalized linear model (GLM) family.

Both DNNs and random effect models have been successful
in improving prediction accuracy of linear models
but in different ways.
Recently, there has been a rising interest in combining these two extensions. 
\citet{simchoni21, simchoni23} proposed the linear mixed model neural network 
for continuous (Gaussian) outputs with Gaussian random effects, 
which allow explicit expressions for likelihoods. 
\citet{lee23} introduced the hierarchical likelihood (h-likelihood) approach, 
as an extension of classical likelihood for Gaussian outputs,
which provides an efficient likelihood-based procedure.
For non-Gaussian (discrete) outputs, 
\citet{tran20} proposed a Bayesian approach for DNNs with normal random effects 
using the variational approximation method \citep{bis06, ble17}. 
\citet{mandel23} used a quasi-likelihood approach \citep{breslow93} for DNNs, 
but the quasi-likelihood method has been criticized for its prediction accuracy. 
\citet{lee01} proposed the use of Laplace approximation
to have approximate maximum likelihood estimators (MLEs). 
Although \citet{mandel23} also applied Laplace approximation for DNNs, 
their method ignored many terms in the second derivatives due to computational expense, 
which could lead to inconsistent estimations \citep{lee17}. 
Therefore, a new approach is desired for non-Gaussian DNNs 
to obtain the exact MLEs for fixed parameters.

Clustered count data are widely encountered 
in various fields \citep{rou07,hen03, tha90, hen98},
which often involve high-cardinality categorical features,
i.e., categorical variables with a large number of
unique levels or categories, such as subject ID or cluster name.
However, to the best of our knowledge, 
there appears to be no available source code 
for the subject-specific Poisson DNN models. 
In this paper, we introduce Poisson-gamma DNN for the clustered count data
and derive the h-likelihood that simultaneously provides 
MLEs of fixed parameters and best unbiased predictors (BUPs) of random effects. 
In contrast to the ordinary HGLM and DNN framework, 
we found that the local minima can cause poor prediction 
when the DNN model contains subject-specific random effects. 
To resolve this issue, we propose an adjustment to the random effect prediction 
that prevents from violation of the constraints for identifiability. 
Additionally, we introduce the method-of-moments estimators 
for pretraining the variance component.
It is worth emphasizing that 
incorporating state-of-the-art network architectures
into the proposed h-likelihood framework is straightforward.
As an example, we implement a feature selection method with multi-head attention.

In Section 2 and 3, we present the Poisson-gamma DNN
and derive its h-likelihood, respectively.
In Section 4, we present the algorithm for online learning, 
which includes an adjustment of random effect predictors, 
pretraining variance components, 
and feature selection method using multi-head attention layer. 
In Section 5, we provide experimental studies 
to compare the proposed method with various existing methods.
The results of the experimental studies clearly show that 
the proposed method improves predictions from existing methods. 
In Section 6, real data analyses demonstrate that 
the proposed method has the best prediction accuracy in various clustered count data. 
Proofs for theoretical results are derived in Appendix.
Source codes are included in Supplementary Materials.

\section{Model Descriptions}

\subsection{Poisson DNN}

Let $y_{ij}$ denote a count output
and $\bx_{ij}$ denote a $p$-dimensional vector of input features, 
where the subscript $(i,j)$ indicates the $j$th outcome 
of the $i$th subject (or cluster) for $i=1,...,n$ and $j=1,...,q_i$.
For count outputs,
Poisson DNN \citep{rod20} gives the marginal predictor,
\begin{equation}
\eta_{ij}^{m}=\log \mu_{ij}^{m}
=\text{NN}(\mathbf{x}_{ij};\mathbf{w},\boldsymbol{\beta})
=\sum_{k=1}^{p_{L}}g_{k}(\mathbf{x}_{ij};\mathbf{w})\beta_{k}+\beta_{0},  
\label{eq:nn}
\end{equation}
where $\mu_{ij}^{m}=\text{E}(Y_{ij}|\mathbf{x}_{ij})$ is the marginal mean,
NN$(\mathbf{x}_{ij};\mathbf{w},\boldsymbol{\beta})$ is the neural network predictor,
$\boldsymbol{\beta}=\left( \beta_{0},\beta_{1},...,\beta_{p_{L}}\right)^{T}$
is the vector of weights and bias between the last
hidden layer and the output layer, $g_{k}(\mathbf{x}_{ij};\mathbf{w})$
denotes the $k$-th node of the last hidden layer, and $\mathbf{w}$ is the
vector of all the weights and biases before the last hidden layer. Here the
inverse of the log function, $\exp (\cdot)$, becomes the activation function of
the output layer. Poisson DNNs allow highly nonlinear relationship between 
input and output variables, but only provide the marginal predictions for $\mu_{ij}^m$.
Thus, Poisson DNN can be viewed as an extension of Poisson GLM with
$\eta_{ij}^{m}=\mathbf{x}_{ij}^T\beta$.

\subsection{Poisson-gamma DNN}

To allow subject-specific prediction into the model \eqref{eq:nn}, 
we propose the Poisson-gamma DNN,
\begin{equation}
\eta_{ij}^{c}=\log \mu_{ij}^{c}
=\text{NN}(\mathbf{x}_{ij};\mathbf{w},\boldsymbol{\beta})
+\mathbf{z}_{ij}^{T}\mathbf{v},  
\label{eq:hgnnm}
\end{equation}
where $\mu_{ij}^{c}=\text{E}(Y_{ij}|\mathbf{x}_{ij},v_{i})$ is the
conditional mean, $\text{NN}(\mathbf{x}_{ij};\mathbf{w},\boldsymbol{\beta})$
is the marginal predictor of the Poisson DNN \eqref{eq:nn},
$\mathbf{v}=(v_{1},...,v_{n})^{T}$ is
the vector of random effects from the log-gamma distribution, 
and $\mathbf{z}_{ij}$ is a vector from the model matrix for random effects,
representing the high-cardinality categorical features. The conditional mean $\mu_{ij}^c$
can be formulated as
\begin{equation*}
\mu_{ij}^{c}=\exp \left\{ 
\text{NN}(\mathbf{x}_{ij};\mathbf{w},\boldsymbol{\beta})\right\} \cdot u_{i},
\end{equation*}
where $u_{i}=\exp (v_{i})$ is the gamma random effect.

\begin{figure}[tbp]
\begin{center}
\includegraphics[width=0.9\linewidth]{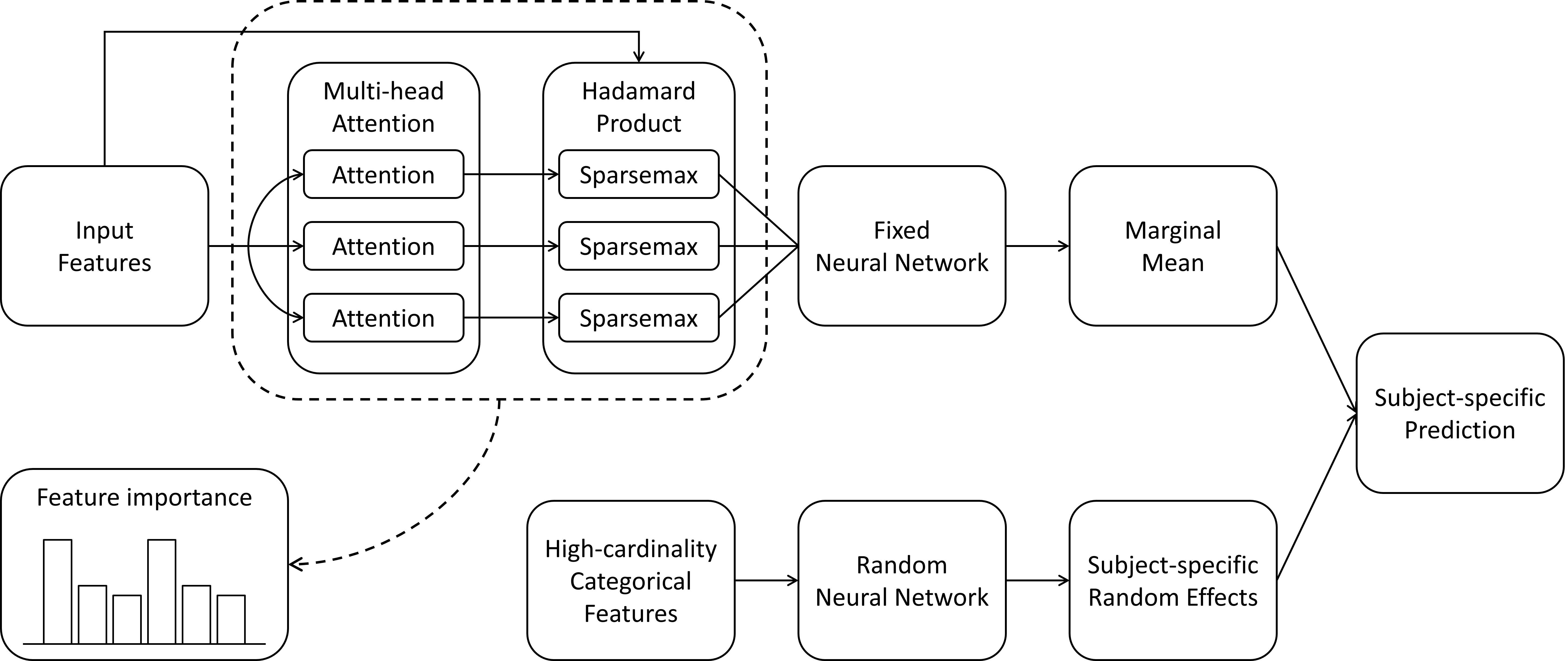}
\end{center}
\caption{An example of the proposed model architecture.
The input features are denoted by $\bx_{ij}$
and the high-cardinality categorical features are denoted by $\bz_{ij}$
in the proposed model \eqref{eq:hgnnm}.}
\label{fig:model}
\end{figure}

Note here that, for any $\epsilon \in \mathbb{R}$, 
the model \eqref{eq:hgnnm} can be expressed as
\begin{equation*}
\log \mu_{ij}^{c}
=\sum_{k=1}^{p_{L}}g_{k}(\mathbf{x}_{ij};\mathbf{w})\beta_{k}+\beta_{0}+v_{i}
=\sum_{k=1}^{p_{L}}g_{k}(\mathbf{x}_{ij};\mathbf{w})\beta_{k}
+(\beta_{0}-\epsilon)+(v_{i}+\epsilon),
\end{equation*}
or equivalently, for any $\delta = \exp(\epsilon) > 0$,
\begin{equation*}
\mu_{ij}^{c}
= \exp \left\{\text{NN}(\mathbf{x}_{ij};\mathbf{w},\boldsymbol{\beta})\right\}
\cdot u_i
= \exp \left\{ 
\text{NN}(\mathbf{x}_{ij};\mathbf{w},\boldsymbol{\beta})
- \log \delta
\right\} \cdot (\delta u_i),
\end{equation*}
which leads to an identifiability problem. Thus, it is necessary to place
constraints on either the fixed parts 
$\text{NN}(\mathbf{x}_{ij};\mathbf{w},\boldsymbol{\beta})$
or the random parts $u_i$.
\citet{lee23} developed subject-specific DNN models 
with Gaussian outputs,
imposing the constraint $\text{E}(v_{i})=0$,
which is common for normal random effects.
For Poisson-gamma DNNs, we use the constraints 
$\text{E}(u_{i})=\text{E}(\exp (v_{i}))=1$ 
for subject-specific prediction of count outputs.
The use of constraints $\text{E}(u_{i})=1$ 
has an advantage that the marginal predictions 
for multiplicative model can be directly obtained, 
because
\begin{equation*}
\mu_{ij}^{m}
=\text{E}[\text{E}(Y_{ij}|\textbf{x}_{ij},u_{i})]
=\text{E}\left[
\exp \left\{ \text{NN}(\mathbf{x}_{ij};\mathbf{w},\boldsymbol{\beta})\right\} 
\cdot u_{i}\right] 
=\exp \left\{ \text{NN}(\mathbf{x}_{ij};\mathbf{w},\boldsymbol{\beta})\right\}.
\end{equation*}
Thus, we employ $v_{i}=\log u_{i}$ to the model \eqref{eq:hgnnm}, where 
$u_{i}\sim \text{Gamma}(\lambda ^{-1},\lambda ^{-1})$ 
with $\text{E}(u_{i})=1$
and $\text{var}(u_{i})=\lambda $. By allowing two separate output nodes, the
Poisson-gamma DNN provides both marginal and subject-specific predictions,
\begin{equation*}
\widehat{\mu}_{ij}^{m}=\exp (\widehat{\eta}_{ij}^{m})
=\exp \left\{ 
\text{NN}(\mathbf{x}_{ij};\widehat{\mathbf{w}},\widehat{\boldsymbol{\beta}})
\right\} 
\quad \text{and} \quad \widehat{\mu}_{ij}^{c}
=\exp \left\{ \text{NN}(\mathbf{x}_{ij};\widehat{\mathbf{w}},\widehat{\boldsymbol{\beta}})
+\mathbf{z}_{ij}^{T}\widehat{\mathbf{v}}\right\},
\end{equation*}
where the hats denote the predicted values. Subject-specific prediction can
be achieved by multiplying the marginal mean predictor 
$\widehat{\mu}_{ij}^{m}$ and the subject-specific predictor of 
random effect $\widehat{u}_{i}=\exp (\widehat{v}_{i})$.
Note here that 
\begin{equation*}
\text{var}(Y|\mathbf{x})
=\text{E}(\text{var}(Y|\mathbf{x},\mathbf{v}))
+\text{var}(\text{E}(Y|\mathbf{x},\mathbf{v}))
\geq \text{E}(\text{var}(Y|\mathbf{x},\mathbf{v})),
\end{equation*}
where $\text{var}(\text{E}(Y|\mathbf{x},\mathbf{v}))$ 
represents the between-subject variance and 
$\text{E}(\text{var}(Y|\mathbf{x},\mathbf{v}))$
represents the within-subject variance. 
To enhance the predictions,
Poisson DNN improves the marginal predictor 
$\bmu^m = \text{E}(Y|\mathbf{x})=$E$\{$E$(Y|\mathbf{x},\mathbf{v})\}$ 
by allowing highly nonlinear function of $\mathbf{x}$,
whereas 
Poisson-gamma DNN further uses the conditional predictor 
$\bmu^c = \text{E}(Y|\mathbf{x},\mathbf{v})$, 
eliminating between-subject variance. 
Figure \ref{fig:model} illustrates an example 
of the proposed model architecture including feature selection
in Section \ref{sec:algorithm}.

\section{Construction of h-likelihood}
\label{sec:hlik}

For subject-specific prediction via random effects,
it is important to define the objective
function for obtaining exact MLEs of fixed parameters 
$\boldsymbol{\theta}=(\mathbf{w},\boldsymbol{\beta},\lambda)$. 
In the context of linear mixed models,
\citet{henderson59} proposed to maximize
the joint density with respect to fixed and random parameters.
However, it cannot yield MLEs of variance components.
There have been various attempts to extend joint maximization schemes
with different justifications
\citep{gilmour85, harville84, schall91, breslow93, wolfinger93},
but failed to obtain simultaneously the exact MLEs of all fixed parameters
and BUPs of random parameters
by optimizing a single objective function.
It is worth emphasizing that defining joint density
requires careful consideration because of the Jacobian term
associated with the random parameters.
For $\boldsymbol{\theta}$ and $\mathbf{u}$, 
an extended likelihood \citep{lee17} can be defined as
\begin{equation}
\ell_{e}(\boldsymbol{\theta},\mathbf{u})
=\sum_{i,j}\log f_{\boldsymbol{\theta}}(y_{ij}|u_{i})
+\sum_{i}\log f_{\boldsymbol{\theta}}(u_{i}).  
\label{eq:extended}
\end{equation}
However, a nonlinear transformation $v_i=v(u_i)$ of random effects $u_i$
leads to different extended likelihood due to the Jacobian terms:
\begin{align*}
\ell_{e}(\boldsymbol{\theta},\mathbf{v})
& = \sum_{i,j}\log f_{\boldsymbol{\theta}}(y_{ij}|v_{i})
+\sum_{i}\log f_{\boldsymbol{\theta}}(v_{i}) 
\\
& = \sum_{i,j}\log f_{\boldsymbol{\theta}}(y_{ij}|u_{i})
+\sum_{i}\log f_{\boldsymbol{\theta}}(u_{i})
+\sum_{i}\log \left| \frac{du_{i}}{dv_{i}}\right|
\neq \ell_{e}(\boldsymbol{\theta},\mathbf{u}).
\end{align*}
The two extended likelihoods 
$\ell_{e}(\boldsymbol{\theta},\mathbf{u})$ and 
$\ell_{e}(\boldsymbol{\theta},\mathbf{v})$ lead to different estimates, 
raising the question on how to obtain the true MLEs. 
In Poisson-gamma HGLMs, \citet{lee96} proposed the use of 
$\ell_{e}(\boldsymbol{\theta},\mathbf{v})$
that can give MLEs for $\boldsymbol{\beta}$ 
and BUPs for $\mathbf{u}$ by the joint maximization.
However, it could not yield MLE for the variance component $\lambda$. 
In this paper, we derive the new h-likelihood 
whose joint maximization simultaneously yields 
MLEs of the whole fixed parameters 
including the variance component $\lambda$,
BUPs of the random effects $\mathbf{u}$,
and conditional expectations $\boldsymbol{\mu}^{c}$.

Suppose that $\bv^* = (v_1^*, ..., v_n^*)^T$ is 
a transformation of $\bv$ such that
$$
v_i^* = v_i \cdot \exp \{ - c_i(\btheta; \by_i) \},
$$
where $c_i(\btheta; \by_i)$ is a function of $\btheta$ 
and $\by_i = (y_{i1}, ..., y_{iq_i})^T$
for $i = 1,2,...,n$.
Then we define the h-likelihood as
\begin{equation}
h(\btheta, \bv) 
\equiv \log f_{\btheta}(\bv^*|\by) + \log f_{\btheta}(\by)
= \ell_e(\btheta, \bv) 
+ \sum_{i=1}^{n} c_i(\btheta, \by_i),
\label{eq:hlik}
\end{equation}
if the joint maximization of $h(\btheta, \bv)$ leads to 
MLEs of all the fixed parameters and BUPs of the random parameters.
A sufficient condition for $h(\boldsymbol{\theta},\mathbf{v})$ 
to yield exact MLEs of all the fixed parameters in $\boldsymbol{\theta}$ 
is that $f_{\boldsymbol{\theta}}(\widetilde{\mathbf{v}}^{*}|\mathbf{y})$ 
is independent of $\boldsymbol{\theta}$, 
where $\widetilde{\mathbf{v}}^{*}$ is the mode, 
\begin{equation*}
\widetilde{\mathbf{v}}^{*}
=\argmax_{\mathbf{v}^{*}}
h(\boldsymbol{\theta},\mathbf{v}^{*})
=\argmax_{\mathbf{v}^{*}}
\log f_{\boldsymbol{\theta}}(\mathbf{v}^{*}|\mathbf{y}).
\end{equation*}
For the proposed model, Poisson-gamma DNN,
we found that the following function
$$
c_{i}(\boldsymbol{\theta};\mathbf{\mathbf{y}}_{i})
=c_{i}(\lambda;y_{i+})
=(y_{i+}+\lambda^{-1})+\log \Gamma (y_{i+}+\lambda^{-1})
-(y_{i+}+\lambda^{-1})\log (y_{i+}+\lambda^{-1})
$$
satisfies the sufficient condition,
\begin{equation*}
\log f(\widetilde{\mathbf{v}}^{*}|\mathbf{y})
=\sum_{i=1}^{n}\log f_{\boldsymbol{\theta}}
(\widetilde{v}_{i}^{*}|\mathbf{y})
=\sum_{i=1}^{n}\left\{ \log f_{\boldsymbol{\theta}}
(\widetilde{v}_{i}|\mathbf{y})
+c_{i}(\boldsymbol{\theta};\mathbf{\mathbf{y}}_{i})\right\} 
=0,
\end{equation*}
where $y_{i+}=\sum_{j=1}^{q_i} y_{ij}$ 
is the sum of outputs in $\mathbf{y}_{i}$ and 
$\widetilde{v}_i = \widetilde{v}^*_i \cdot \exp\{ c_i(\btheta; \by_i) \}$.
Then, the h-likelihood at mode 
$h(\boldsymbol{\theta},\widetilde{\mathbf{v}})$
becomes the classical (marginal) log-likelihood,
\begin{equation}
\ell (\boldsymbol{\theta};\mathbf{y})
=\log f_{\btheta}(\by)
=\log \int \exp \left\{ 
\ell_{e}(\boldsymbol{\theta},\mathbf{v})
\right\} d\mathbf{v}.
\label{eq:marginal}
\end{equation}
Thus, joint maximization of the h-likelihood \eqref{eq:hlik} 
provides exact MLEs for the fixed parameters $\boldsymbol{\theta}$, 
including the variance component $\lambda $. 
BUPs of $\mathbf{u}$ and $\boldsymbol{\mu}^{c}$ 
can be also obtained from our h-likelihood,
\begin{equation*}
\widetilde{\mathbf{u}}=\exp (\widetilde{\mathbf{v}})
=\text{E}(\mathbf{u}|\mathbf{y})
\quad \text{and}\quad 
\widetilde{\boldsymbol{\mu}}^{c}
=\exp (\widetilde{\mathbf{v}}) \cdot 
\exp \left\{ \text{NN}(\mathbf{X};\mathbf{w},\boldsymbol{\beta}) \right\}
=\text{E}(\boldsymbol{\mu}^{c}|\mathbf{y}).
\end{equation*}
The proof and technical details for the theoretical results 
are derived in Appendix \ref{app:hlik}.

\section{Learning algorithm for Poisson-gamma DNN models}
\label{sec:algorithm}
In this section, we introduce the h-likelihood learning framework
for handling the count data with high-cardinality categorical features.
We decompose the negative h-likelihood loss for online learning
The entire learning algorithm of the proposed method 
is briefly described in Algorithm \ref{alg:hl}.

\begin{algorithm}[tb]
   \caption{Learning algorithm for Poisson-gamma DNN via h-likelihood}
   \label{alg:hl}
\begin{algorithmic}
\STATE
   \STATE {\bfseries Input:} $\bx_{ij}$, $\bz_{ij}$
   \FOR{epoch $=0$ {\bfseries to} pretrain epochs}
   \STATE Train $\bw$, $\bbeta$ and $\bv$ by minimizing the negative h-likelihood.
   \STATE Compute method-of-moments estimator of $\lambda$.
   \STATE Adjust the random effect predictors.
   \ENDFOR
   \FOR{epoch $=0$ {\bfseries to} train epochs}
   \STATE Train all the fixed and random parameters by minimizing the negative h-likelihood.
   \STATE Adjust the random effect predictors.
   \ENDFOR
   \STATE Compute MLE of $\lambda$.
\end{algorithmic}
\end{algorithm}

\subsection{Loss function for online learning}

The proposed Poisson-gamma DNN can be trained by optimizing the negative
h-likelihood loss, 
\begin{align*}
\text{Loss}& =-h(\boldsymbol{\theta},\mathbf{v})
=-\log f_{\boldsymbol{\theta}}(\mathbf{y}|\mathbf{v})
-\log f_{\boldsymbol{\theta}}(\mathbf{v})-c(\boldsymbol{\theta};\mathbf{y}),
\end{align*}
which is a function of the two separate output nodes 
$\mu_{ij}^{m}=\text{NN}(\mathbf{x}_{ij};\mathbf{w},\boldsymbol{\beta})$ 
and $v_{i}=\mathbf{z}_{ij}^{T}\mathbf{v}$. 
To apply online stochastic optimization methods, 
the proposed loss function is expressed as 
\begin{equation}
\text{Loss}
=\sum_{i,j}\left[ -y_{ij}\left( \log \mu_{ij}^{m}+v_{i}\right)
+e^{v_{i}}\mu_{ij}^{m}-\frac{v_{i}-e^{v_{i}}}{q_{i}\lambda}
+a_{i}(\lambda ;\mathbf{y}_{i})\right],  
\label{eq:loss}
\end{equation}
where $a_{i}(\lambda ;\mathbf{y}_{i})
=q_{i}^{-1}\left\{ \lambda ^{-1}\log\lambda 
+\log \Gamma (\lambda ^{-1})-c_{i}(\lambda,y_{i+})\right\}.$

\subsection{Random Effect Adjustment}

While DNNs often encounter local minima, 
\citet{dauphin14} claimed that in ordinary DNNs, 
local minima may not necessarily result in poor predictions.
In contrast to HGLM and DNN, 
we observed that the local minima can lead to poor prediction 
when the network reflects subject-specific random effects. 
In Poisson-gamma DNNs, 
we impose the constraint $\text{E}(u_{i})=1$ for identifiability, 
because for any $\delta>0$,
\begin{equation*}
\mu_{ij}^{c}
= \exp \left\{ 
\text{NN}(\mathbf{x}_{ij}; \mathbf{w},\boldsymbol{\beta})
\right\} \cdot u_{i} 
= \left[ \exp \left\{
\text{NN}(\mathbf{x}_{ij}; \mathbf{w},\boldsymbol{\beta}) 
-\log \delta
\right\} \right] \cdot \left( \delta u_i \right).
\end{equation*}
However, in practice, Poisson-gamma DNNs often end with local minima 
that violate the constraint. 
To prevent poor prediction due to local minima,
we introduce an adjustment to the predictors of $u_{i}$, 
\begin{equation}
\widehat{u}_{i}\leftarrow 
\frac{\widehat{u}_{i}}{\frac{1}{n}
\sum_{i=1}^{n}\widehat{u}_{i}}
\quad \text{and}\quad 
\widehat{\beta}_{0}\leftarrow 
\widehat{\beta}_{0}+\log \left( \frac{1}{n}
\sum_{i=1}^{n}\widehat{u}_{i}\right)  
\label{eq:adjust}
\end{equation}
to satisfy $\sum_{i=1}^{n}\widehat{u}_{i}/n=1$. 
The following theorem shows that the proposed adjustment 
improves the local h-likelihood prediction.
The proof is given in Appendix \ref{app:thm}.

\begin{theorem}
In Poisson-gamma DNNs, suppose that $\widehat{\beta}_{0}$ 
and $\widehat{u}_{i}$ are estimates of $\beta_{0}$ 
and $u_{i}$ such that 
$
\sum_{i=1}^{n}\widehat{u}_{i}/n=1+\epsilon 
$
for some $\epsilon \in \mathbb{R}$. 
Let $\widehat{u}_{i}^{*}$ and $\widehat{\beta}_{0}^{*}$ 
be the adjusted estimators in \eqref{eq:adjust}. Then, 
\begin{equation*}
h(\widehat{\boldsymbol{\theta}}^{*},\widehat{\mathbf{v}}^{*})
\geq h(\widehat{\boldsymbol{\theta}},\widehat{\mathbf{v}}),
\end{equation*}
and the equality holds if and only if $\epsilon =0$, 
where $\widehat{\boldsymbol{\theta}}$ and $\widehat{\boldsymbol{\theta}}^{*}$ 
are vectors of the same fixed parameter estimates 
but with different $\widehat{\beta}_{0}$ 
and $\widehat{\beta}_{0}^{*}$ for $\beta_{0}$,
respectively.
\end{theorem}

\begin{figure}[tbp]
\begin{center}
\includegraphics[width=0.9\linewidth]{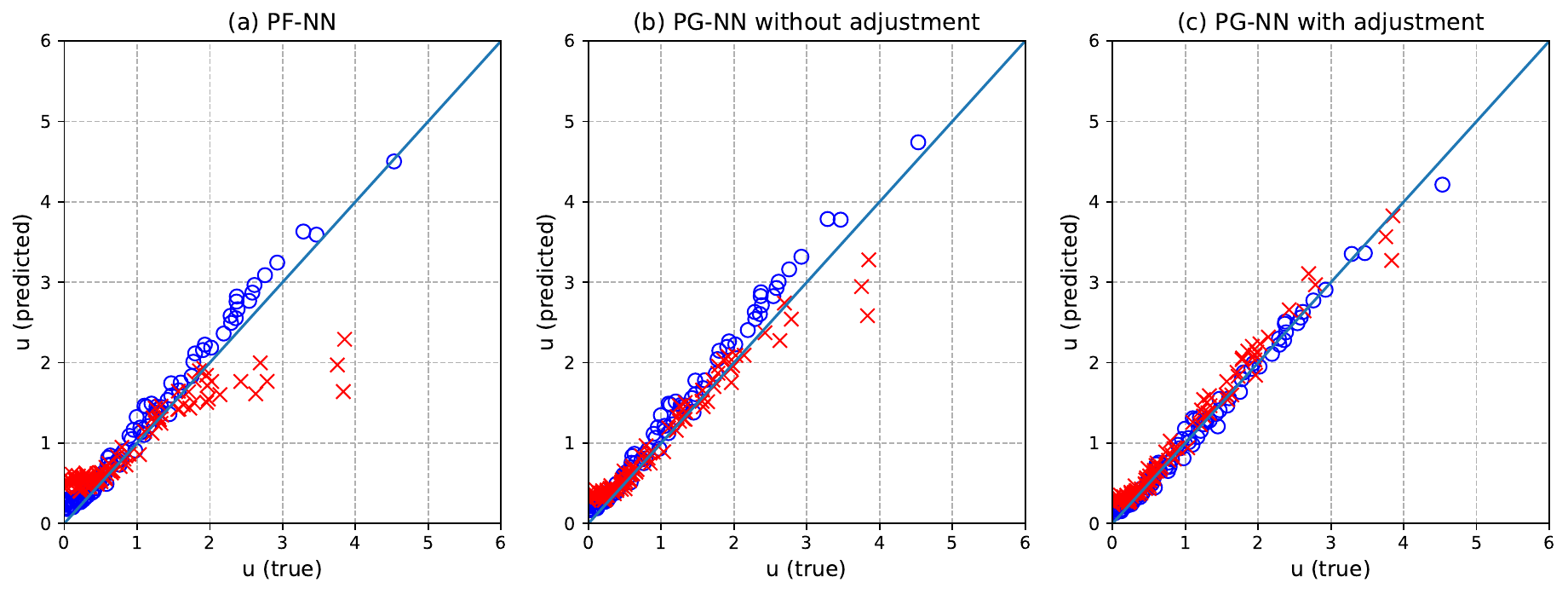}
\end{center}
\caption{Predicted values of $u_{i}$ from two replications (marked as o and
x for each) when $u_{i}$ is generated from the Gamma distribution with 
$\lambda = 1$, $n=100$, $q=100$.}
\label{fig:uhat}
\end{figure}

Theorem 1 shows that the adjustment \eqref{eq:adjust} 
improves the random effect prediction.
According to our experience, even though limited, this
adjustment becomes important, especially when the cluster size is large.
Figure \ref{fig:uhat} is the plot of $\widehat{u}_{i}$ 
against the true $u_{i}$ under $(n,q)=(100,100)$ and $\lambda = 1$. 
Figure \ref{fig:uhat} (a) shows that 
the use of fixed effects for subject-specific effects (PF-NN) 
produces poor prediction of $u_i$.
Figure \ref{fig:uhat} (b) and (c) show that 
the use of random effects for subject-specific effects (PG-NN) 
improves the subject-specific prediction,
and the proposed adjustment improves it further.

\subsection{Pretraining variance components}

We found that the MLE for variance component $\lambda =\text{var}(u_{i})$ 
could be sensitive to the choice of initial value, giving a slow convergence. 
We propose the use of method-of-moments estimator (MME) for pretraining $\lambda$, 
\begin{equation}
\widehat{\lambda}
=\left[ \frac{1}{n}\sum_{i=1}^{n}(\widehat{u}_{i}-1)^{2}\right] 
\left[ \frac{1}{2}+\sqrt{\frac{1}{4}
+\frac{n\sum_{i}^{n}\widehat{\mu}_{i+}^{-1}(\widehat{u}_{i}-1)^{2}}
{\left\{ \sum_{i}^{n}(\widehat{u}_{i}-1)^{2}\right\} ^{2}}}\right],
\label{eq:lambda estimate}
\end{equation}
where $\widehat{\mu}_{i+} = \sum_{j=1}^{q_i} \widehat{\mu}^m_{ij}$.
Convergence of the MME \eqref{eq:lambda estimate} is
shown in Appendix \ref{app:mme}. Figure \ref{fig:lambda} shows that 
the proposed pretraining accelerates the convergence in various settings. 
In Appendix \ref{app:consistency}, 
we demonstrate an additional experiments for verifying 
the consistency of $\widehat{\lambda}$ of the proposed method.

\begin{figure}[tbp]
\begin{center}
\includegraphics[width=0.9\linewidth]{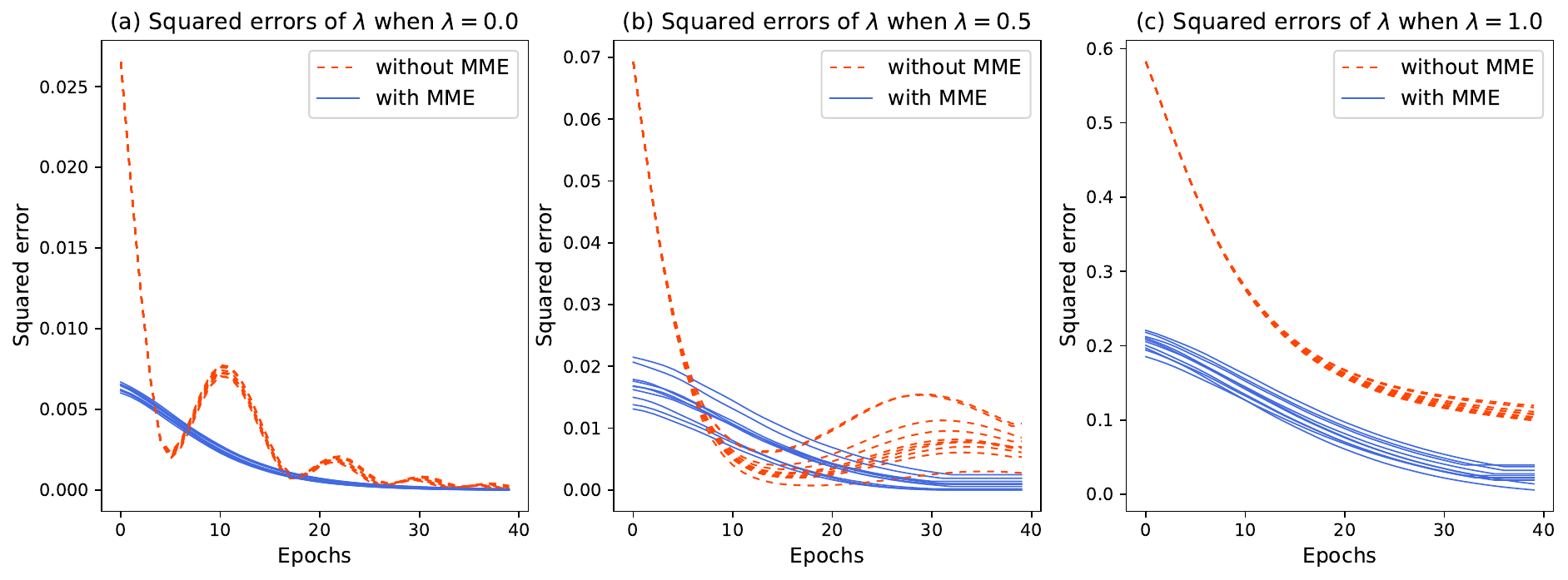}
\end{center}
\caption{Learning curve for the variance component $\lambda$ 
when (a) $\lambda=0$, (b) $\lambda=0.5$, and (c) $\lambda=1$.}
\label{fig:lambda}
\end{figure}

\subsection{Feature Selection in High Dimensional Settings}
\begin{figure}[tbp]
\begin{center}
\includegraphics[width=0.9\linewidth]{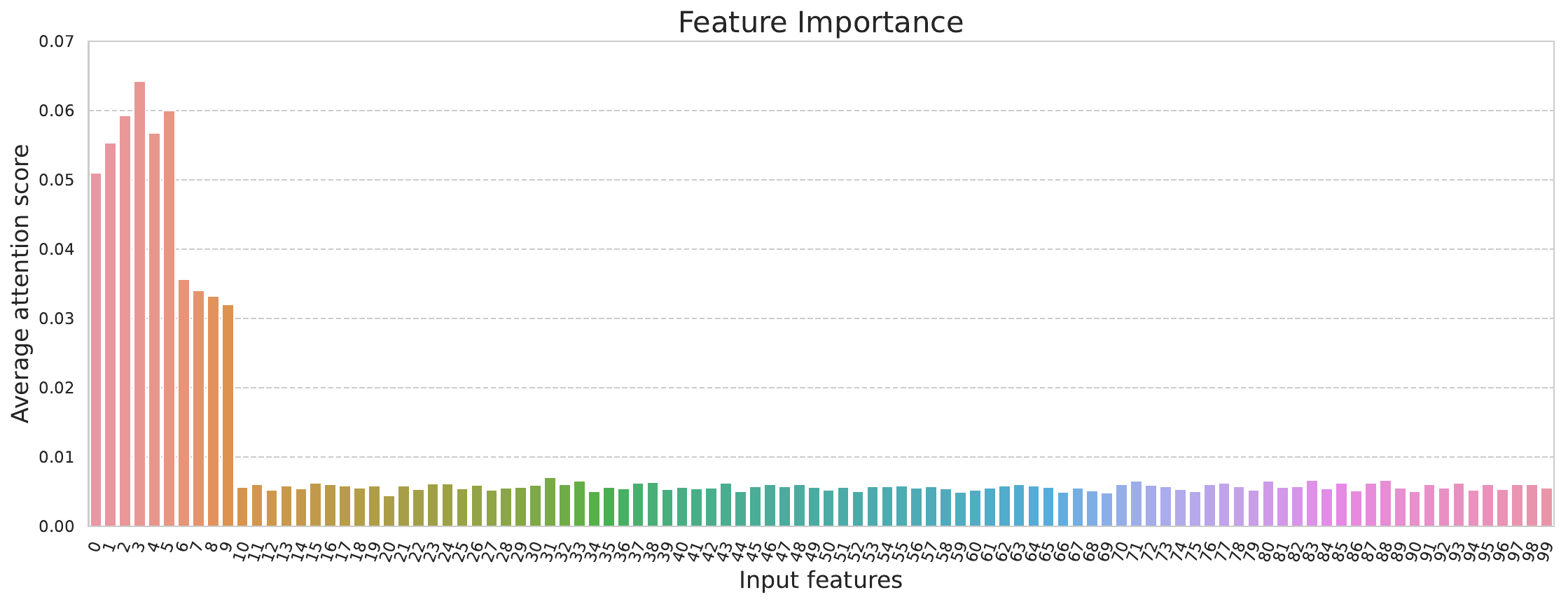}
\end{center}
\caption{Average attention scores. 
First 10 features are genuine and the others are irrelevant.}
\label{fig:attention}
\end{figure}

Feature selection methods can be easily implemented to the proposed PG-NN. 
As an example, we implemented feature selection 
using the multi-head attention layer with sparsemax function
\citep{mar16, vsk20, ari21},
$$
\text{sparsemax}(\bz)
= \argmin_{p \in \Delta^{K-1}} || \bp - \bz ||^2,
$$
where $\Delta^{K-1} = \{ \bp \in \mathbb{R}^{K} : \bone^T \bp = 1, \bp \geq 0 \}$.
As a high dimensional setting, 
we generate input features $x_{kij}$ from $N(0,1)$ for $k=1,...,100$,
including 10 genuine features $(k\leq 10)$ 
and 90 irrelevant features $(k>10)$.
The output $y_{ij}$ is generated from $\text{Poi}(\mu^c_{ij})$
with the mean model,
$$
\mu^c_{ij} = u_i \cdot \exp \left[
0.2 \left\{ 1 + \cos x_{1ij} + \cdots + \cos x_{6ij}
+ (x^2_{7ij}+1)^{-1} + \cdots + (x^2_{10ij}+1)^{-1}
\right\} \right],
$$
where $u_i=e^{v_i}$ is generated from $\text{Gamma}(2,2)$.
The number of subjects is $n=10^4$, 
i.e. the cardinality of the categorical feature is $10^4$.
The number of repeated measures (cluster size) is set to be $q=20$,
which is smaller than number of features $p=100$.
We employed a multi-layer perceptron with 20-10-10 number of nodes 
and three-head attention layer for feature selection.
Other details are derived in Section \ref{sec:experiment}.
Figure \ref{fig:attention} shows the average attention scores of 50 repetitions. 
It is evident that all the genuine features are ranked at the top 10.

\section{Experimental Studies}
\label{sec:experiment}

To investigate the performance of the Poisson-gamma DNN, we conducted
experimental studies. The five input variables 
$\mathbf{x}_{ij}=(x_{1ij},...,x_{5ij})^{T}$ 
are generated from the AR(1) process with autocorrelation $\rho =0.5$ 
for each $i=1,...,n$ and $j=1,...,q$. 
The random effects are generated from either 
$u_{i}\sim \text{Gamma}(\lambda^{-1},\lambda ^{-1})$ 
or $v_{i}\sim \text{N}(0,\lambda)$ where $v_{i}=\log u_{i}$. 
When $\lambda =0$, the conditional mean $\mu_{ij}^{c}$ 
is identical to the marginal mean $\mu_{ij}^{m}$. 
The output variable $y_{ij}$ is generated from Poisson$(\mu^c_{ij})$ with
$$
\mu^c_{ij} = u_i \cdot \exp \left[
0.2 \left\{ 1 + \cos x_{1ij} + \cos x_{2ij} + \cos x_{3ij}
+ (x^2_{4ij}+1)^{-1} + (x^2_{5ij}+1)^{-1}
\right\} \right],
$$
Results are based on the 100 sets of simulated data. 
The data consist of $q=10$ observations for $n=1000$ subjects. 
For each subject, 6 observations are assigned to the training set, 
2 are assigned to the validation set, 
and the remaining 2 are assigned to the test set. 

For comparison, we consider the following models.
\begin{itemize}
\item  \makebox[1.6cm]{\textbf{P-GLM} \hfill}
Classic Poisson GLM for count outputs using \texttt{R}.
\item  \makebox[1.6cm]{\textbf{N-NN} \hfill} 
Conventional DNN for continuous outputs.
\item  \makebox[1.6cm]{\textbf{P-NN} \hfill} 
Poisson DNN for count outputs. 
\item  \makebox[1.6cm]{\textbf{PN-GLM} \hfill} 
Poisson-normal HGLM using \texttt{lme4} \citep{lme4} package in \texttt{R}.
\item  \makebox[1.6cm]{\textbf{PG-GLM} \hfill} 
Poisson-gamma HGLM using the proposed method.
\item  \makebox[1.6cm]{\textbf{NF-NN} \hfill} 
Conventional DNN with fixed subject-specific effects for continuous outputs.
\item  \makebox[1.6cm]{\textbf{NN-NN} \hfill} 
DNN with normal random effects for continuous outputs \citep{lee23}.
\item  \makebox[1.6cm]{\textbf{PF-NN} \hfill} 
Conventional Poisson DNN with fixed subject-specific effects for count outputs.
\item  \makebox[1.6cm]{\textbf{PG-NN} \hfill} 
The proposed Poisson-gamma DNN for count outputs.
\end{itemize}
To evaluate the prediction performances, 
we consider the root mean squared Pearson error (RMSPE)
$$
\text{RMSPE} = \sqrt{\frac{1}{N} \sum_{i,j} 
\frac{(y_{ij}-\widehat{\mu}_{ij})^2}{V(\widehat{\mu}_{ij})}},
$$
where $\text{Var}(y_{ij}|u_i) = \phi V(\mu_{ij})$ 
and $\phi$ is a dispersion parameter of GLM family.
For Gaussian outputs, the RMSPE is identical 
to the ordinary root mean squared error,
since $\phi = \sigma^2$ and $V(\mu_{ij})=1$.
For Poisson outputs, since $\phi = 1$ and $V(\mu_{ij})=\mu_{ij}$,
the RMSPE for test set is given by
\begin{equation*}
\label{eq:rmsp}
\text{RMSPE} = \sqrt{\frac{1}{N_{\text{test}}}\sum_{(i,j)\in \text{test}}
\frac{(y_{ij}-\widehat{\mu}_{ij})^{2}}{\widehat{\mu}_{ij}}}.
\end{equation*}
P-GLM, N-NN, and P-NN give marginal predictions
$\widehat{\mu}_{ij}=\widehat{\mu}_{ij}^m$, 
while the others give subject-specific predictions
$\widehat{\mu}_{ij}=\widehat{\mu}_{ij}^c$. 
N-NN, NF-NN, and NN-NN are models for continuous outputs, 
while the others are models for count outputs. 
For NF-NN and PF-NN, predictions are made by 
maximizing the conditional likelihood 
$\sum_{i,j}\log f_{\boldsymbol{\theta}}(y_{ij}|v_{i})$. 
On the other hand, for PN-GLM, PG-GLM, NN-NN, and PG-NN, 
subject-specific predictions are made by maximizing the h-likelihood.

PN-GLM is the generalized linear mixed model 
with random effects $v_{i}\sim N(0,\lambda)$. 
Current statistical software for PN-GLM and PG-GLM 
(\texttt{lme4} and \texttt{dhglm})
provide approximate MLEs using Laplace approximation.
The proposed learning algorithm can yield exact MLEs for PG-GLM, 
using solely the input and output layers while excluding the hidden layers.
Among various methods for NN-NN 
\citep{tran20, mandel23, simchoni21, simchoni23, lee23}, 
we applied the state-of-the-art method proposed by \citet{lee23}. 
All the DNNs and PG-GLMs were implemented in Python 
using Keras \citep{keras} and TensorFlow \citep{tensorflow}.
For all DNNs, we employed a standard multi-layer perceptron (MLP) 
consisting of 3 hidden layers with 10 neurons 
and leaky ReLU activation function. 
We applied the Adam optimizer with a learning rate of 0.001 
and an early stopping process based on the validation loss 
while training the DNNs. 
NVIDIA Quadro RTX 6000 were used for computations.

\begin{table}
\caption{
Mean and standard deviation of test RMSPEs of simulation studies
over 100 replications.
G(0) implies the absence of random effects,
i.e., $v_i=0$ for all $i$.
Bold numbers indicate the minimum.}
\label{tb:rmsp_simul}
\begin{center}
\begin{tabular}{cccccc}
\toprule
                & \multicolumn{5}{c}{Distribution of random effects ($\lambda$)} \\ \cmidrule{2-6}
Model           &G(0) \& N(0)          &G(0.5)      &G(1)          &N(0.5)      &N(1)      \\ \midrule 
P-GLM    &1.046 (0.029)    &1.501 (0.055)     &1.845 (0.085)       &1.745 (0.113)       &2.818 (0.467) \\
N-NN     &1.013 (0.018)    &1.473 (0.042)     &1.816 (0.074)       &1.713 (0.097)       &1.143 (0.432) \\
P-NN     &\bf1.011 (0.018) &1.470 (0.042)     &1.812 (0.066)       &1.711 (0.099)       &1.161 (0.440) \\
PN-GLM   &1.048 (0.029)    &1.112 (0.033)	  &1.115 (0.035)       &1.124 (0.030)	    &1.152 (0.034) \\
PG-GLM   &1.048 (0.020)    &1.123 (0.027)     &1.106 (0.023)       &1.139 (0.026)       &1.161 (0.028) \\
NF-NN    &1.152 (0.029)    &1.301 (0.584)     &1.136 (0.311)       &1.241 (1.272)       &1.402 (0.298) \\
NN-NN    &1.020 (0.020)    &1.121 (0.026)     &1.209 (0.067)       &1.256 (0.097)       &2.773 (0.384) \\
PF-NN    &1.147 (0.025)    &1.135 (0.029)     &1.128 (0.027)       &1.129 (0.024)       &1.128 (0.027) \\
PG-NN    &1.016 (0.019)    &\bf1.079 (0.024)  &\bf1.084 (0.023)    &\bf1.061 (0.022)    &\bf1.085 (0.026) \\
\bottomrule
\end{tabular}
\end{center}
\end{table}
Table \ref{tb:rmsp_simul} shows the mean and standard deviation of test RMSPEs 
from the experimental studies. 
When the true model does not have random effects (G(0) and N(0)), 
the PG-NN is comparable to the P-NN without random effects, 
which should perform the best (marked by the bold face) in terms of RMSPE. 
N-NN (P-NN) without random effects is also better than 
NF-NN and NN-NN (PF-NN and PG-NN) with random effects. 
When the distribution of random effects is correctly specified (G(0.5) and G(1)), 
the PG-NN performs the best in terms of RMSPE. 
Even when the distribution of random effects is misspecified (N(0.5), N(1)), 
the PG-NN still performs the best. 
This result is in accordance with the simulation results of \citet{mcc11}, 
namely, in GLMMs, the prediction accuracy is little affected 
for violations of the distributional assumption for random effects:
see similar performances of PN-GLM and PG-GLM.

It has been known that handling the high-cardinality categorical features as random effects 
has advantages over handling them as fixed effects \citep{lee17}, 
especially when the cardinality of categorical feature is close to the sample size,
i.e., the number of observations in each category (cluster size $q$) is relatively small. 
Thus, to emphasize the advantages of PG-NN over PF-NN
in high-cardinality categorical features, 
we consider two additional scenarios for experimental study with cluster size 
$q_{\text{train}}=3$ and $q_{\text{train}}=1$ where $\lambda=0.2$ and $n=1000$.
Mean and standard deviation of RMSPE of PF-NN are 1.269 (0.038) and 1.629 (0.086)
for $q_{\text{train}}=3$ and $q_{\text{train}}=1$, respectively.
Those of PG-NN are 1.124 (0.028) and 1.284 (0.049) for each scenario.
Therefore, the proposed method enhances subject-specific predictions 
as the cardinality of categorical features becomes high.

\section{Real Data Analysis}

\begin{table}
\caption{
Test RMSPEs of real data analyses.
Bold numbers indicate the minimum values.}
\label{tb:rmsp_real}
\begin{center}
\begin{tabular}{cccccc}
\toprule
        &\multicolumn{5}{c}{Dataset} \\ \cmidrule{2-6}
Model   &Epilepsy	   &CD4	       &Bolus      &Owls       &Fruits     \\ \midrule 
P-GLM   &1.520         &6.115      &2.110      &2.307      &6.818      \\ 
N-NN    &2.119         &8.516      &1.982      &2.297      &6.573      \\ 
P-NN    &1.712         &6.830      &2.354      &3.076      &6.854      \\ 
PN-GLM  &1.242         &3.422      &1.727      &5.791      &6.795      \\ 
PG-GLM  &1.229         &4.424      &1.714      &4.479      &\bf5.786   \\ 
NF-NN   &1.750         &6.921      &1.718      &2.215      &5.897      \\ 
NN-NN   &1.770         &7.640      &1.727      &2.674      &5.825      \\ 
PF-NN   &1.238         &3.558      &1.816      &2.951      &6.430      \\ 
PG-NN   &\bf1.135      &\bf3.513   &\bf1.677   &\bf2.000   &6.376      \\ \bottomrule
\end{tabular}
\end{center}
\end{table}

To investigate the prediction performance of clustered count outputs in practice, 
we examined the following five real datasets:

\begin{itemize}
\item  \textbf{Epilepsy data} \citep{tha90} 
\item  \textbf{CD4 data} \citep{hen98}
\item  \textbf{Bolus data} \citep{hen03}
\item  \textbf{Owls data} \citep{rou07} 
\item  \textbf{Fruits data} \citep{ban10} 
\end{itemize}

For all the DNNs, a standard MLP with one hidden layer of 10 neurons and
a sigmoid activation function were employed. 
For longitudinal data (Epilepsy, CD4, Bolus), 
the last observation for each patient was used as the test set. 
For clustered data (Owls, Fruits), 
an observation was randomly selected as the test set from each cluster. 
RMSPEs are reported in Table \ref{tb:rmsp_real},
which shows that the use of subject-specific models (PG-NN and PG-GLM) 
for count data are the best.
Throughout the datasets, P-GLM performs better than P-NN,
implying that non-linear model does not improve the linear model
in the absence of subject-specific random effects.
Meanwhile, in the presence of subject-specific random effects,
PG-NN is always preferred to PG-GLM except for Fruits data.
The results imply that introducing subject-specific random effects in DNNs 
can help to identify the nonlinear effects of the input variables.
Therefore, while DNNs are widely recognized for improving predictions 
in independent datasets, 
introducing subject-specific random effects could be necessary for DNNs 
to improve their predictions in correlated datasets
with high-cardinality categorical features.

\section{Concluding Remarks}
When the data contains high-cardinality categorical features, 
introducing random effects into DNNs is advantageous.
We develop subject-specific Poisson-gamma DNN for clustered count data.
The h-likelihood enables a fast end-to-end learning algorithm
using the single objective function.
By introducing subject-specific random effects, 
DNNs can effectively identify the nonlinear effects of the input variables. 
Various state-of-the-art network architectures can be easily implemented
into the h-likelihood framework, 
as we demonstrate with the feature selection based on multi-head attention.

\bibliography{references}
\bibliographystyle{apalike}

\appendix
\section{Appendix}
\subsection{Derivation of the h-likelihood}
\label{app:hlik}
Maximizing $\log f_{\btheta}(\bv|\by)$ 
with respect to $\bv = (v_1, ..., v_n)^T$ yields,
\begin{align*}
\tilde{v}_i
\equiv \argmax_{v_i} \log f_{\boldsymbol{\theta}}(v_i|\mathbf{y}) 
= \argmax_{v_i} \left[	\sum_{j=1}^{q_i} 
\left( y_{ij} v_i - \mu_{ij}^{m} e^{v_i} \right) 
+ \frac{ v_i - e^{v_i}}{\lambda}  \right]
= \log \left( \frac{y_{i+} +\lambda^{-1}}{\mu_{i+} +\lambda^{-1}} \right),
\end{align*}
where $y_{i+} = \sum_{j=1}^{q_i} y_{ij}$ 
and $\mu_{i+} = \sum_{j=1}^{q_i} \mu^{m}_{ij}$.
As derived in Section \ref{sec:hlik}, 
define $c_i(\lambda; y_{i+})$ as
$$
c_{i}(\lambda;y_{i+})
=(y_{i+}+\lambda^{-1})+\log \Gamma (y_{i+}+\lambda^{-1})
-(y_{i+}+\lambda^{-1})\log (y_{i+}+\lambda^{-1})
$$
and consider a transformation,
$$
v_i^* = v_i \cdot \exp \{ - c_i(\btheta; \by_i) \}.
$$
Since the multiplier $\exp \{ - c_i(\btheta; \by_i) \}$
does not depend on $v_i$, we have
$$
\tilde{v}^*_i
\equiv \argmax_{v^*_i} \log f_{\boldsymbol{\theta}}(v^*_i|\mathbf{y}) 
= \tilde{v}_i \cdot \exp \{ - c_i(\btheta; \by_i) \},
$$
which leads to
$$
\log f(\widetilde{\bv}^*|\by)
= \sum_{i=1}^n \left\{
\log f_{\btheta}(\widetilde{v}_i|\by) + c_i(\btheta; \by_i)
\right\}
= 0.
$$
This satisfies the sufficient condition for the h-likelihood
to give exact MLEs for fixed parameters,
$$
\argmax_{\btheta} h(\btheta, \widetilde{\bv})
= \argmax_{\btheta} \ell(\btheta;\by).
$$
Furthermore, from the distribution of $u_i|\by_i$,
$$
\widetilde{u}_i = \exp(\widetilde{v}_i)
= \frac{y_{i+} +\lambda^{-1}}{\mu_{i+} +\lambda^{-1}}
= \text{E} (u_i | \by_i)
$$
and
$$
\widetilde{\mu}_{ij}^c 
= \exp(\widetilde{v}_i) \cdot 
\exp \left\{ \text{NN}(\bX; \bw, \bbeta) \right\}
= \mu_{ij}^m \cdot \text{E}(u_i|\by_i)
= \text{E}(\mu_{ij}^c|\by).
$$
Therefore, maximizing the h-likelihood yields
the BUPs of $u_i$ and $\mu_{ij}^c$.

\subsection{Proof of Theorem 1.}
\label{app:thm}
The adjustment \eqref{eq:adjust} transports
$$
\widehat{u}_i^* = \widehat{u}_i/(1+\epsilon)
\quad \text{and} \quad
\widehat{v}_i^* = \widehat{v}_i - \log(1+\epsilon).
$$
Since $(\widehat{\boldsymbol{\theta}}, \widehat{\mathbf{v}})$ and $( \widehat{\boldsymbol{\theta}}^*, \widehat{\mathbf{v}}^*)$
have the same conditional expectation $\widehat{\mu}_{ij}$, equation \eqref{eq:hlik} yields
\begin{align*}
h ( \widehat{\boldsymbol{\theta}}^*, \widehat{\mathbf{v}}^*) 
- h ( \widehat{\boldsymbol{\theta}}, \widehat{\mathbf{v}})
&= \sum_{i=1}^{n}\left[ \frac{ \widehat{v}_{i}^*
- \exp(\widehat{v}_{i}^*)}{\widehat{\lambda}} \right]
- \sum_{i=1}^{n}\left[ \frac{ \widehat{v}_{i}
- \exp(\widehat{v}_{i})}{\widehat{\lambda}} \right]
\\
&= \widehat{\lambda}^{-1} \left\{
	\sum_{i=1}^{n} \left( \widehat{v}_{i}^* - \widehat{v}_{i} \right)
	- \sum_{i=1}^{n} \left( \widehat{u}_{i}^* - \widehat{u}_{i} \right)
\right\}
\\
&= n \widehat{\lambda}^{-1}\left\{\epsilon - \log (1+\epsilon)\right\}
\geq 0,
\end{align*}
and the equality holds if and only if $\epsilon = 0$.
Thus, $h( \widehat{\boldsymbol{\theta}}^*, \widehat{\mathbf{v}}^*) 
\geq h ( \widehat{\boldsymbol{\theta}}, \widehat{\mathbf{v}})$.

\subsection{Convergence of the method-of-moments estimator}
\label{app:mme}
As derived in Section A, for given $\lambda$ and $\mu_{i+}$, 
maximization of the h-likelihood leads to
$$
\widehat{u}_i = \widehat{u}_i (\mathbf{y}_i) 
= \exp \left( \widehat{v}_i (\mathbf{y}_i) \right) 
= \frac{y_{i+}+\lambda^{-1}}{\mu_{i+}+\lambda^{-1}}.
$$
Thus, $\text{E}(\widehat{u}_i)=1$ and 
$\text{Var}(\widehat{u}_i) 
= \lambda \left\{ 1 - (\lambda \mu_{i+} + 1)^{-1} \right\}$.
Define $d_i$ as
\begin{align*}
d_i &= \frac{\widehat{u}_i - 1}{\sqrt{1 - (\lambda \mu_{i+} + 1)^{-1}}}
= \frac{y_{i+}-\mu_{i+}}{\mu_{i+}+\lambda^{-1}} \sqrt{1 
+ \lambda^{-1} \mu_{i+}^{-1}}
\end{align*}
to have $\text{E}(d_i)=0$ and $\text{Var}(d_i)=\lambda$ for any $i=1,...,n$. 
Then, by the law of large numbers,
$$
\frac{1}{n} \sum_{i=1}^n d_i^2 
\to \text{E}(d_i^2) = \text{Var}(d_i) + \text{E}(d_i)^2 
= \lambda.
$$
Note here that
$$
\frac{1}{n} \sum_{i=1}^n d_i^2
= \left\{ \frac{1}{n} \sum_{i=1}^n (\widehat{u}_i-1)^2 \right\}
+ \frac{1}{\lambda} \left\{\frac{1}{n} 
\sum_{i=1}^n \frac{(\widehat{u}_i-1)^2}{\mu_{i+}} \right\}.
$$
Then, solving the following equation,
$$
\lambda
- \left\{ \frac{1}{n} \sum_{i=1}^n (\widehat{u}_i-1)^2 \right\}
- \frac{1}{\lambda} \left\{\frac{1}{n} 
\sum_{i=1}^n \frac{(\widehat{u}_i-1)^2}{\widehat{\mu}_{i+}} \right\}
= 0,
$$
leads to an estimate $\widehat{\lambda}$ in \eqref{eq:lambda estimate}
and $\widehat{\lambda} \to \lambda$ as $n\to \infty$.

\subsection{Consistency of variance component estimator}
\label{app:consistency}
\begin{figure}[tbp]
\begin{center}
\includegraphics[width=0.5\linewidth]{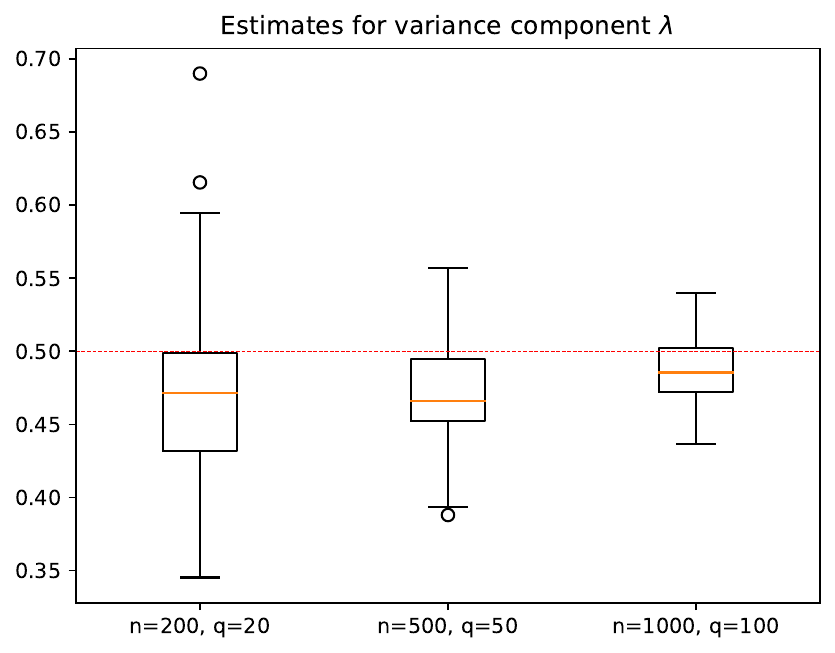}
\end{center}
\caption{Box-plots of $\widehat{\lambda}$
when true $\lambda = 0.5$.}
\label{fig:consistency}
\end{figure}

By jointly maximizing the proposed h-likelihood, 
we obtain MLEs for fixed parameters and BUPs for random parameters. 
Thus, we can directly apply the consistency of MLEs under usual regularity conditions. 
To verify this consistency in PG-NN, 
we present the boxplots in Figure \ref{fig:consistency}
for the variance component estimator $\widehat{\lambda}$
from 100 repetitions with true $\lambda=0.5$. 
The number of clusters $n$ and cluster size $q$ 
are set to be (200, 20), (500, 50), and (1000,100). 
Figure \ref{fig:consistency} provide conclusive evidence 
confirming the consistency of the variance component estimator 
within the PG-NN framework.

\subsection{Details of the real datasets}
\label{app:data}
\begin{itemize}
\item  \textbf{Epilepsy data:} Epilepsy data are reported by \citet{tha90} 
from a clinical trial of $n=59$ patients with epilepsy. 
The data contain $N=236$ observations with $q_{i}=4$ repeated measures from each patient 
and $p=4$ input variables. 

\item  \textbf{CD4 data:} CD4 data are from a study of AIDS patients with
advanced immune suppression, reported by \citet{hen98}. 
The data contain $N=4612$ observations from $n=1036$ patients with 
$q_{i}\geq 2$ repeated measurements and $p=4$ input variables. 

\item  \textbf{Bolus data:} Bolus data are from a clinical trial following
abdominal surgery for $n=65$ patients with $q_i=12$ repeated measurements,
reported in \citet{hen03}. The data have $N=780$ observations with $p=2$ input variables.

\item  \textbf{Owls data:} Owls data are reported by \citet{rou07}, 
which can be found in the \texttt{R} package \texttt{glmmTMB} \citep{glmmTMB}. 
The data contain $N=599$ observations and $n=27$ nests with $p=3$ input variables.
The cluster size $q_{i}$ in each nest varies from 4 to 52.

\item  \textbf{Fruits data:} Fruits data are reported in \citet{ban10}. 
The data have $N=625$ observations clustered by $n=24$ types of maternal seed
family with $p=3$ input variables. The cluster size $q_{i}$ varies from 11 to 47.
\end{itemize}

\end{document}